\PassOptionsToPackage{table}{xcolor}
\documentclass[fleqn,10pt]{wlscirep}
\usepackage[utf8]{inputenc}
\usepackage[T1]{fontenc}
\usepackage{float}
\usepackage{caption}
\usepackage{amsmath}
\usepackage{hyperref}
\usepackage{lettrine}
\usepackage{longtable}
\usepackage{booktabs}
\usepackage{geometry}
\usepackage{tabularx}
\usepackage{arydshln}
\definecolor{myred}{rgb}{1.0, 0.0, 0.0}
\definecolor{myblue}{rgb}{0.0, 0.0, 1.0}
\DeclareCaptionLabelFormat{adja-page}{}

\title{\LARGE \textbf{SCOPE-DTI}: \textmd{\textbf{S}emi-Inductive Dataset \textbf{C}onstruction and Framework \textbf{O}ptimization for \textbf{P}ractical Usability \textbf{E}nhancement in Deep Learning-Based \textbf{D}rug \textbf{T}arget \textbf{I}nteraction Prediction}}

\author[1, 2, 5, †]{Yigang Chen}
\author[1, 2, †]{Xiang Ji}
\author[1, 2, †]{Ziyue Zhang}
\author[1]{Yuming Zhou}
\author[1, 2, 3, 5]{Yang-Chi-Dung Lin}
\author[1, 2, 3]{Hsi-Yuan Huang}
\author[5]{Tao Zhang}
\author[1, 2]{Yi Lai}
\author[1, 2]{Ke Chen}
\author[1]{Chang Su}
\author[1]{Xingqiao Lin}
\author[1, 2]{Zihao Zhu}
\author[1]{Yanggyi Zhang}
\author[1]{Kangping Wei}
\author[1]{Jiehui Fu}
\author[1, 2]{Yixian Huang}
\author[1, 2]{Shidong Cui}
\author[1, 2]{Shih-Chung Yen}
\author[6]{Ariel Warshel}
\author[1, 2, 3, 4, 5, *]{Hsien-Da Huang}

\affil[1]{School of Medicine, The Chinese University of Hong Kong, Shenzhen, Longgang District, Shenzhen, Guangdong 518172, China}
\affil[2]{Warshel Institute for Computational Biology, School of Medicine, The Chinese University of Hong Kong, Shenzhen, Longgang District, Shenzhen, Guangdong 518172, China}
\affil[3]{Guangdong Provincial Key Laboratory of Digital Biology and Drug Development, The Chinese University of Hong Kong, Shenzhen, Longgang District, Shenzhen, Guangdong 518172, China}
\affil[4]{Department of Endocrinology, Key Laboratory of Endocrinology of National Health Commission, Peking Union Medical College Hospital, Chinese Academy of Medical Sciences \& Peking Union Medical College, Beijing, 100730, P.R. China}
\affil[5]{Better Way Group - Chinese University of Hong Kong (Shenzhen) Warshel Joint Laboratory for skin health and active molecule innovation, Longgang District, Shenzhen, Guangdong 518172, China}
\affil[6]{Department of Chemistry, University of Southern California, Los Angeles, California 90089-1062, United States}
\affil[ ]{\textsuperscript{†}These authors contributed equally to this work.}
\affil[*]{corresponding author: Hsien-Da Huang (huanghsienda@cuhk.edu.cn)}

\usepackage{lineno}
\linenumbers

\begin{abstract}
Deep learning-based drug-target interaction (DTI) prediction methods have demonstrated strong performance; however, real-world applicability remains constrained by limited data diversity and modeling complexity. To address these challenges, we propose SCOPE-DTI, a unified framework combining a large-scale, balanced semi-inductive human DTI dataset with advanced deep learning modeling. Constructed from 13 public repositories, the SCOPE dataset expands data volume by up to 100-fold compared to common benchmarks such as the Human dataset. The SCOPE model integrates three-dimensional protein and compound representations, graph neural networks, and bilinear attention mechanisms to effectively capture cross domain interaction patterns, significantly outperforming state-of-the-art methods across various DTI prediction tasks. Additionally, SCOPE-DTI provides a user-friendly interface and database. We further validate its effectiveness by experimentally identifying anticancer targets of Ginsenoside Rh1. By offering comprehensive data, advanced modeling, and accessible tools, SCOPE-DTI accelerates drug discovery research.
\end{abstract}
\begin{document}

\flushbottom
\nolinenumbers
\maketitle

\thispagestyle{empty}

\section*{Introduction}

Identifying drug–target interactions (DTIs) is essential for drug discovery, drug repurposing, toxicity prediction, and improving the success rate of clinical trials \cite{luoNetworkIntegrationApproach2017,pushpakomDrugRepurposingProgress2019,mizutaniRelatingDrugproteinInteraction2012,smitSystematicAnalysisProtein2021}. For decades, experimental approaches have dominated DTI identification due to their high accuracy and reliability \cite{hajareReviewHighthroughputScreening2013, frimanMassSpectrometrybasedCellular2020}. However, their high cost and extensive time requirements limit scalability, especially considering the immense chemical and biological search spaces \cite{huangRobustDrugTargetInteraction2023}.

The accumulation of extensive experimental datasets has enabled the development of data-driven computational methods \cite{koutsoukasSilicoTargetPrediction2011,chenArtificialIntelligenceDrug2023}. Machine learning-based approaches leverage these large-scale datasets to learn and generalize interaction patterns, facilitating rapid and cost-effective prediction of novel drug–target interactions (DTIs). These computational models provide efficient predictions, guiding experimental validation toward the most promising candidate interactions, thereby accelerating drug discovery and significantly reducing associated costs \cite{bagherianMachineLearningApproaches2021}.

Recent advances in deep learning have significantly advanced computational DTI prediction. Proteochemometrics (PCM), a widely-used computational framework, represents drugs and targets as feature vectors to predict DTIs through supervised classification \cite{vanwestenProteochemometricModelingTool2011}. Embedding methods for molecules and proteins have rapidly evolved, shifting from initial one-dimensional convolutional neural network-based methods \cite{ozturkDeepDTADeepDrugtarget2018,leeDeepConvDTIPredictionDrugtarget2019} to transformer-based models like BERT \cite{huangRobustDrugTargetInteraction2023,karimiDeepAffinityInterpretableDeep2019}, and subsequently to graph neural networks (GNNs) \cite{baiInterpretableBilinearAttention2023,tsubakiCompoundproteinInteractionPrediction2019} and three-dimensional spatial embeddings \cite{luoCalibratedGeometricDeep2023,zhengPredictingDrugProtein2020}. In parallel, model architectures have evolved from simple multilayer perceptrons (MLPs) to interactive mechanisms like interaction maps \cite{huangMolTransMolecularInteraction2021} and bilinear attention networks (BANs) \cite{baiInterpretableBilinearAttention2023}. These innovations have enabled state-of-the-art models to achieve notable performance improvements on standard benchmarks, including BindingDB \cite{gilsonBindingDB2015Public2016,baiHierarchicalClusteringSplit2021}, human \cite{liuImprovingCompoundproteinInteraction2015}, and KIBA datasets \cite{tangMakingSenseLargescale2014}.

However, substantial challenges remain when applying these models in practical settings. In practical drug discovery applications, accurately predicting interactions between novel compounds and druggable targets is essential. Therefore, models must demonstrate the ability to achieve reliable predictive performance even when operating beyond the limitations of their training data. Chatterjee et al. \cite{chatterjeeImprovingGeneralizabilityProteinligand2023} categorize drug-target interaction (DTI) prediction into three distinct scenarios: transductive, semi-inductive, and inductive. The inductive scenario, where both drugs and proteins are absent from the training set, is crucial for novel DTI discovery but remains highly challenging. Current state-of-the-art models under inductive conditions achieve only moderate predictive accuracy, with AUROC typically below 0.7 and correlation coefficients below 0.5 \cite{baiInterpretableBilinearAttention2023,luoCalibratedGeometricDeep2023}, limiting their practical utility. Conversely, the transductive scenario, where both drugs and proteins are present in the training set but their interactions are unknown, yields high performance but is often compromised by overfitting and dataset biases \cite{chenTransformerCPIImprovingCompoundprotein2020,baiHierarchicalClusteringSplit2021}, thus restricting generalization to real-world tasks.

In contrast, the semi-inductive scenario—particularly the prediction of interactions involving new drugs against known proteins—offers significant potential compared to purely inductive or transductive approaches \cite{luoCalibratedGeometricDeep2023,chatterjeeImprovingGeneralizabilityProteinligand2023}. Although current models perform well in such scenarios, their success remains limited by the size and diversity of available data. Expanding and integrating existing datasets to encompass more druggable targets could substantially enhance predictive accuracy and real-world applicability, enlightening the practical value of the semi-inductive approach.

To overcome these limitations, we propose SCOPE-DTI (Semi-Inductive Dataset Construction and Framework Optimization for Practical Usability Enhancement in Deep Learning-Based Drug Target Interaction Prediction), a unified framework designed to enhance the practical utility of DTI prediction models. To our knowledge, we constructed the largest semi-inductive, well balanced and human-focused DTI dataset, integrating data from 13 public repositories and expanding available training data volume by 20 to 100 times compared to standard benchmarks \cite{gilsonBindingDB2015Public2016,baiHierarchicalClusteringSplit2021}. Utilizing this dataset, the SCOPE model incorporates three-dimensional structural embeddings, graph neural networks, and a bilinear attention mechanism, achieving superior predictive performance compared to existing state-of-the-art methods across diverse prediction tasks. Additionally, we developed a user-friendly web interface and searchable database to facilitate easy access and usability. The practical effectiveness of SCOPE-DTI was further demonstrated by experimentally validating novel anticancer targets of Ginsenoside Rh1, highlighting the model’s real-world applicability.

In summary, our contributions are as follows: (1) the construction of a largest, well balanced dataset for semi-inductive DTI prediction, (2) the optimization of a prediction framework for enhanced performance, (3) the development of a user-friendly web interface for accessing the dataset and utilizing the model, and (4) the identification and validation of anticancer targets of Ginsenoside Rh1, showcasing real-world applicability in drug discovery.

\begin{figure*}[!t]
    \begin{center}
    \includegraphics[width=0.95\textwidth]{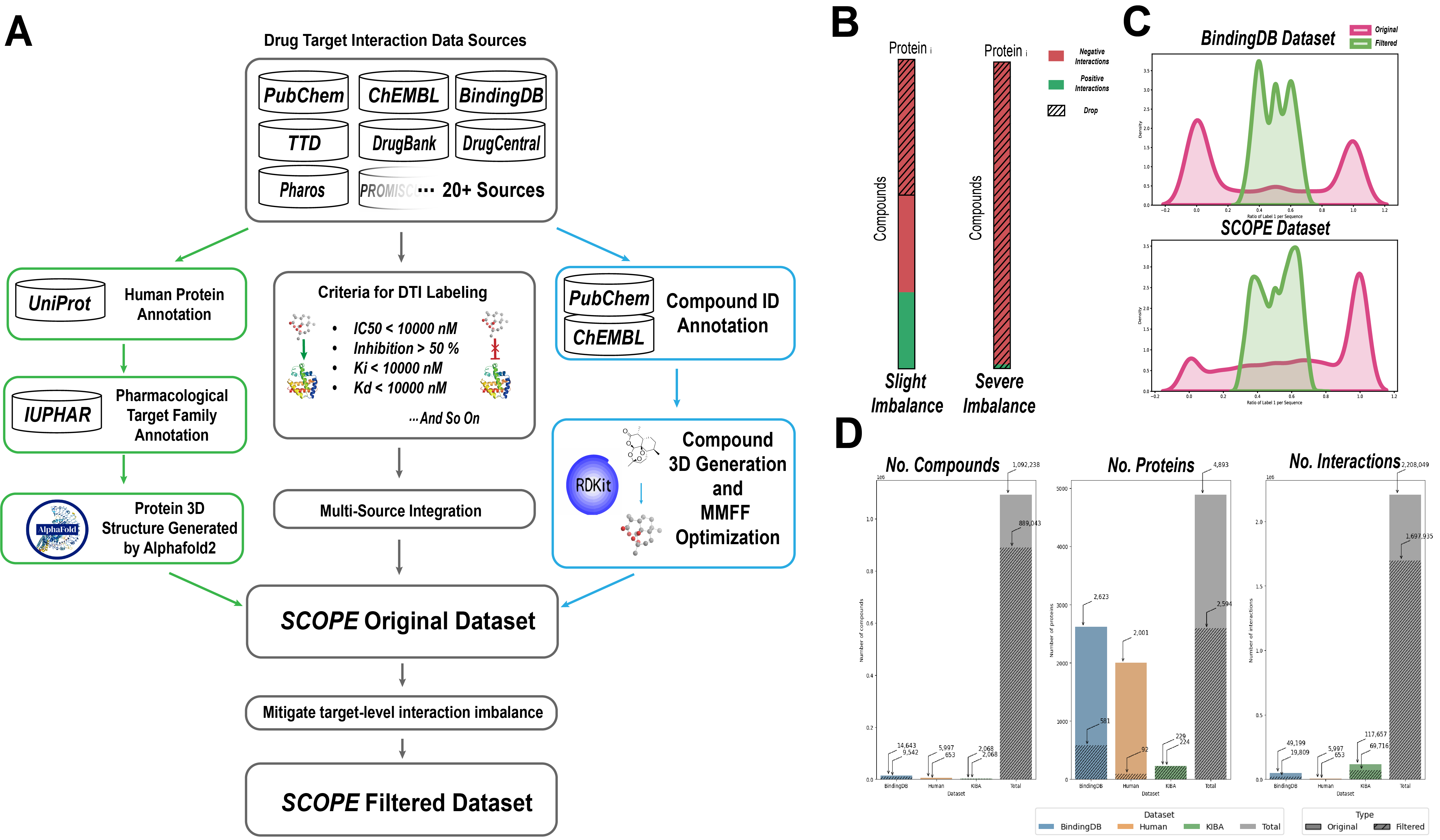}
    \end{center}
    \vspace{-2em}
    \caption{\textbf{Overview of the SCOPE dataset construction pipeline and data characteristics.} \textbf{(A)} Schematic representation of the data integration and preprocessing steps used to construct the SCOPE dataset. \textbf{(B)} Strategy to mitigate target-level imbalance: for proteins dominated by one class, we randomly remove interactions to achieve balance; in extreme cases, we discard all of that protein's interactions. \textbf{(C)} The effect of data filtering. The red and green distributions represent datasets before and after filtering, respectively. \textbf{(D)} Comparison of data volume between SCOPE and previous datasets in terms of the number of compounds, targets, and interactions separately.}
    \label{fig:data_processing}
\end{figure*}

\section*{Results}
\subsection*{SCOPE Dataset Development}
Our study aims to predict drug-target interactions (DTIs) for novel compounds within a semi-inductive framework, addressing the critical question: \textit{Given any novel active compound, how can we accurately identify its binding targets across a comprehensive set of druggable proteins?} To tackle this challenge, we constructed a large-scale, high-quality human-focused DTI dataset.

As illustrated in Figure \ref{fig:data_processing}A, we aggregated and curated data from 13 primary DTI repositories, integrating information from over 20 sources in total, including referenced datasets (Supplementary Table 1). Each dataset entry comprises a protein, a compound, and an interaction label. Proteins were annotated using UniProt identifiers \cite{uniprotconsortiumUniProtUniversalProtein2023}, retaining only human proteins, and categorized into pharmacological families (e.g., GPCRs, kinases) according to the IUPHAR database \cite{hardingIUPHARBPSGuide2024}. To enable structural modeling, we generated 3D protein structures using AlphaFold2 \cite{jumperHighlyAccurateProtein2021}. Compounds were annotated with identifiers from PubChem \cite{kimPubChem2023Update2023} and ChEMBL \cite{zdrazilChEMBLDatabase20232024}, and their 3D structures were constructed using RDKit \cite{RDKitOpensourceCheminformatics} and optimized with the Merck Molecular Force Field (MMFF) \cite{toscoBringingMMFFForce2014}. Interaction labels were systematically assigned using standardized measurement-specific cutoff values derived from PubChem Bioactivity data, classifying interactions as positive (1) or negative (0) (Supplementary Figure 1; Supplementary Table 2). Additional details on labeling methods are described in Supplementary Note.

Recognizing that target-level class imbalance—where certain proteins predominantly display interactions from one class—could bias semi-inductive predictions, we implemented a filtering approach (Figure \ref{fig:data_processing}B). For proteins whose interactions exceeded 75\% from a single class, we randomly removed interactions using stratified sampling, balancing the classes within a 50–75\% range. Proteins with insufficient interactions or extremely skewed class distributions were entirely excluded to maintain dataset integrity. Figure \ref{fig:data_processing}C illustrates that this filtering approach significantly mitigated imbalance, ensuring a more balanced distribution of interaction labels across targets (Supplementary Figure 2 provides further details).

Finally, we quantitatively compared the resulting SCOPE dataset with widely-used benchmarks (Figure \ref{fig:data_processing}D), revealing a substantial increase in scale—20- to 100-fold more interactions, accompanied by significantly more proteins and compounds. Detailed dataset statistics are summarized in Supplementary Table 3. Overall, to our knowledge, SCOPE represents the largest, most balanced semi-inductive human DTI dataset available, providing a robust foundation for developing and validating predictive models for novel drug-target interactions.

\begin{figure*}[!t]
    \begin{center}
    \includegraphics[width=0.9\textwidth]{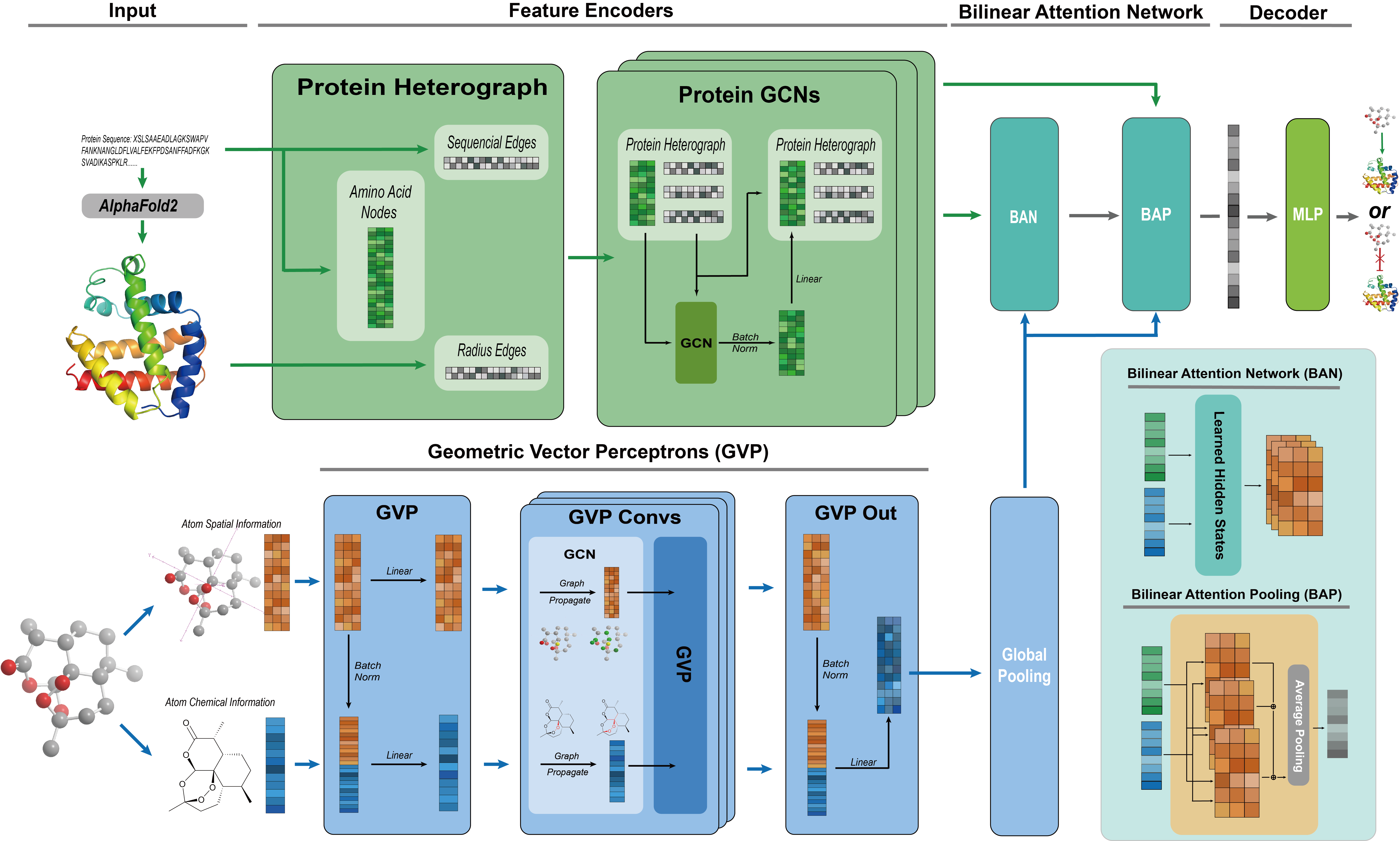}
    \end{center}
    \vspace{-2em}
    \caption{\textbf{Schematic representation of the SCOPE model for drug-target interaction prediction.} The SCOPE model integrates 3D structural information of proteins and compounds to predict drug-target interactions. To capture intricate structural features, the 3D protein structures and molecular graphs are encoded using heterogeneous graph neural networks (HGNNs) and geometric vector perceptrons (GVPs) with global pooling. These encoded representations are then processed through a bilinear attention network, which consists of a bilinear attention layer followed by bilinear pooling, to generate a joint representation that models local interactions between the drug and the target protein. The final predictive score is computed by a fully connected classification layer, representing the likelihood of an interaction.}
    \label{fig:model}
\end{figure*}

\subsection*{SCOPE framework}

Figure \ref{fig:model} provides a schematic illustration of the proposed SCOPE framework for predicting drug-target interactions. The framework begins with the transformation of drug molecules and target proteins from the SCOPE dataset into their respective three-dimensional (3D) conformations, generated using RDKit for compounds and AlphaFold2 for proteins. To effectively capture the complex structural features of both proteins and compounds, their 3D structures are encoded using heterogeneous graph neural networks (HGNNs) \cite{zhangHeterogeneousGraphNeural2019,wuMAPEPPIEffectiveEfficient2024} and geometric vector perceptrons (GVPs) \cite{luoCalibratedGeometricDeep2023,jingLearningProteinStructure2021}, respectively. The protein encoding module simultaneously encodes the sequential relationships among residues and their spatial neighborhood information captured via radius graphs constructed from residue coordinates. The compound encoding incorporates atom-specific features (atom type, charge, hybridization state) defined by DGL-LifeSci’s encoding scheme \cite{liDGLLifeSciOpenSourceToolkit2021}, along with spatial coordinates representing their 3D structural information. Further details of these encoding methods are provided in Supplementary Table 4.

To account for the scale disparity between small molecules and large proteins, atom-level features were aggregated via global pooling to generate fixed-size molecular representations for downstream processing. These pooled molecular vectors are then combined with the encoded protein representations and fed into a BAN. This network comprises a bilinear attention layer, followed by a bilinear pooling layer, enabling the model to capture cross domain interactions between the drug and the target protein residues \cite{baiInterpretableBilinearAttention2023}. Ultimately, the combined representations are passed through a MLP classifier, which computes the final predictive score, representing the probability of a drug-target interaction.

\subsection*{Evaluation strategies and metrics}
We evaluated the classification performance of our model using three publicly available datasets—BindingDB \cite{gilsonBindingDB2015Public2016,baiHierarchicalClusteringSplit2021}, human \cite{liuImprovingCompoundproteinInteraction2015}, and KIBA \cite{tangMakingSenseLargescale2014}—as well as our proprietary SCOPE dataset. To simulate the prediction of DTIs for novel compounds within a semi-inductive framework, we partitioned each dataset by randomly selecting compounds and assigning them to training, validation, and test sets in a 7:1:2 ratio, ensuring that all interactions associated with these compounds were included in the respective sets. Importantly, we ensured that all proteins in the validation and test sets were also represented in the training set, thereby preserving the integrity of the semi-inductive learning setup, where the model is trained on known proteins but evaluated on novel compounds. Detailed information about the dataset splits is provided in Supplementary Note.

To assess the impact of our filtering technique on model performance, we conducted experiments using both the original and filtered versions of all datasets. This comparison enabled us to evaluate the effect of mitigating target-level interaction imbalance on prediction accuracy.

We utilized the area under the receiver operating characteristic curve (AUROC) and the area under the precision-recall curve (AUPRC) as the primary metrics for assessing classification performance. Additionally, we reported accuracy, sensitivity, and specificity at the threshold corresponding to the optimal F1 score. To ensure robustness, we performed at least five independent runs with different random seeds for each dataset split. The model that achieved the highest AUROC on the validation set was selected for final evaluation on the test set, and the performance metrics were then recorded.

\begin{figure*}[!t]
    \begin{center}
    \includegraphics[width=0.9\textwidth]{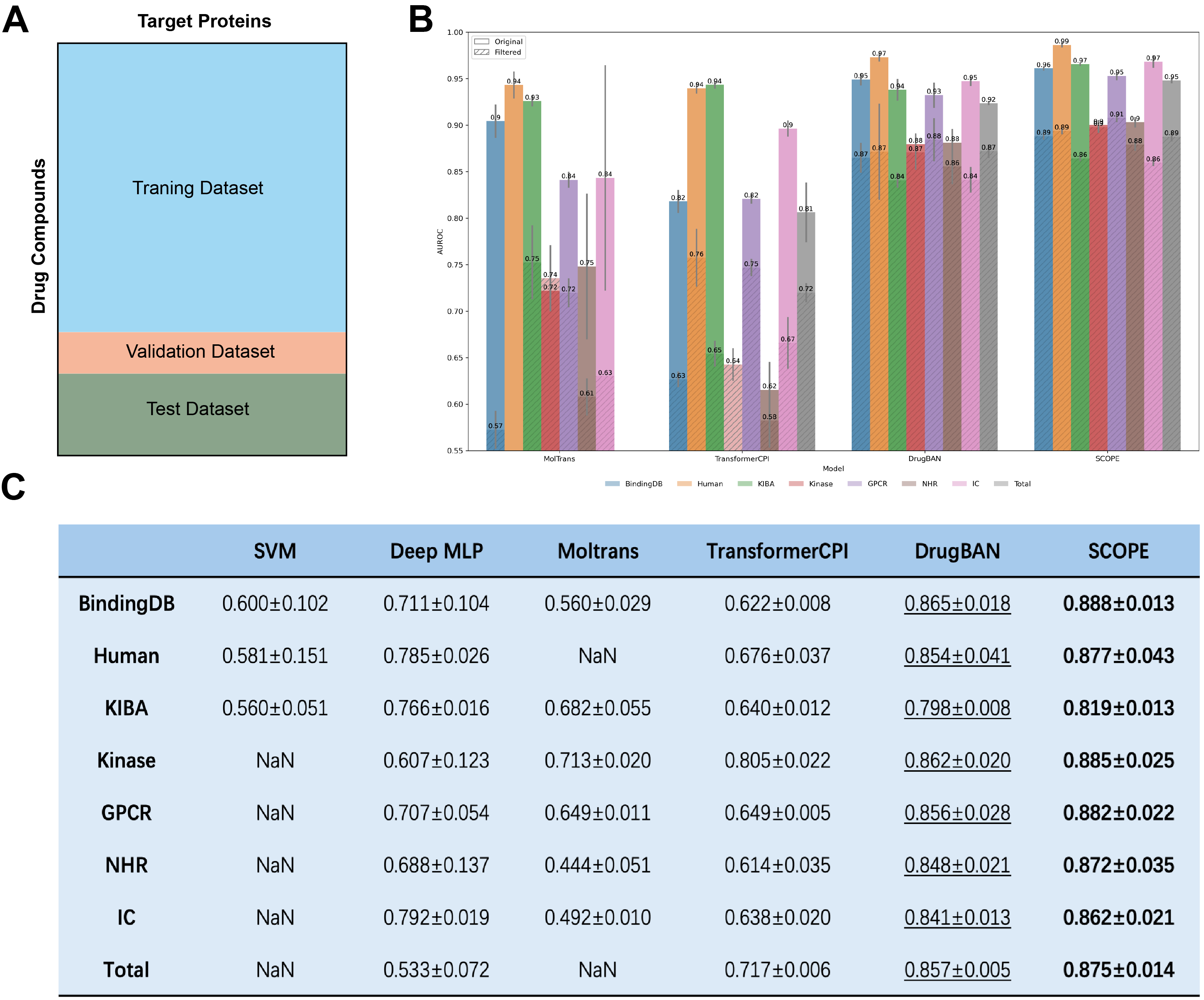}
    \end{center}
    \vspace{-0.5em}
    \caption{\textbf{SCOPE demonstrates superior performance compared to baseline models on a semi-inductive data splitting strategy.} (A) Illustration of the semi-inductive split strategy: drugs are kept unseen in the validation and test sets, while targets remain known from the training dataset. (B) AUPRC performance of four models on eight datasets, before and after dataset filtering. (C) Average AUROC performance of SCOPE and five baseline models across eight filtered datasets. Each experiment was conducted at least five times, and the mean values are presented. SCOPE consistently outperforms all baseline models on every dataset.}
    \label{fig:performance}
\end{figure*}
\subsection*{Performance comparison}

We evaluated the SCOPE model against five baseline models across eight datasets, both before and after applying the bias filtering technique, using a semi-inductive new compound split strategy (Figure~\ref{fig:performance}A). The baseline models included a classical machine learning model, the Support Vector Machine (SVM) \cite{cortesSupportvectorNetworks1995}, a traditional deep learning model, the Multi-Layer Perceptron (MLP) \cite{rumelhartLearningRepresentationsBackpropagating1986}, and three state-of-the-art deep learning models for DTI prediction: MolTrans \cite{huangMolTransMolecularInteraction2021}, TransformerCPI \cite{chenTransformerCPIImprovingCompoundprotein2020}, and DrugBAN \cite{baiInterpretableBilinearAttention2023}. Across all datasets and conditions, SCOPE consistently achieved the highest performance, surpassing the second-best model, DrugBAN, by 1-3\%, and demonstrating significantly greater improvements over the remaining models (Figure~\ref{fig:performance}C). Notably, as detailed in Supplementary Table 5, SCOPE consistently outperformed all baseline models in terms of accuracy, sensitivity, and specificity. Importantly, on the filtered datasets, SCOPE maintained a distinct performance advantage over most models, further demonstrating the robustness and reliability of the approach.

Figure~\ref{fig:performance}B illustrates that the original datasets contained significant bias, which allowed certain models to achieve artificially inflated performance. This bias, particularly evident in smaller datasets such as Human and BindingDB, arises from proteins with limited interactions predominantly belonging to a single interaction class, enabling models to predict outcomes largely based on protein identity alone—an unrealistic scenario in practical drug discovery. Our filtering approach effectively mitigated this bias, generating datasets that more accurately reflect real-world conditions (see Figure~\ref{fig:data_processing}B).

The effectiveness of the filtering strategy is underscored by the substantial performance drop (20–30\% in AUROC and AUPRC) observed in several baseline models, including SVM, MolTrans, TransformerCPI, and the deep MLP, upon evaluation on the filtered datasets. TransformerCPI, in particular, demonstrated significant dependency on the biased characteristics present in the original BindingDB, Human, and overall SCOPE datasets. In contrast, DrugBAN and SCOPE exhibited only minor performance declines after filtering, highlighting their robustness and superior adaptability to realistic, balanced datasets. The consistency of these performance trends further confirms the resilience of DrugBAN and SCOPE architectures under diverse and practical data conditions.

Figure~\ref{fig:performance}C highlights the robust and stable performance of DrugBAN and SCOPE on large-scale datasets. Specifically, the SCOPE Total dataset—both before and after filtering—contains nearly 100 times more interactions than the combined Human and BindingDB datasets. Moreover, the kinase-specific subset of the SCOPE dataset is approximately 20 times larger than the family-specific KIBA dataset. Such significant differences in dataset size provide a clear evaluation of each model's capability to adapt to large-scale data.

Among the baseline models, distinct differences were observed in their capacity to handle large datasets. SVM and MolTrans struggled to scale effectively; their computational designs often exceeded standard hardware limitations when processing large-scale data. Although the deep MLP was capable of fitting the larger datasets, it failed to achieve substantial performance improvements and occasionally exhibited decreased accuracy. In contrast, TransformerCPI, DrugBAN, and SCOPE demonstrated effective scalability and maintained stable performance on larger datasets. Notably, SCOPE consistently achieved the best results, outperforming DrugBAN by 0.02–0.03 AUROC, indicating its superior adaptability and predictive capability on large-scale datasets.

In summary, the superior performance demonstrated by the SCOPE model highlights the significant advantages of data-driven representation learning, especially the integration of 3D structural features for both proteins and compounds. By effectively capturing complex pairwise interaction patterns through its semi-inductive framework, SCOPE achieves enhanced accuracy and robustness in DTI prediction tasks. The integration of high-dimensional, context-sensitive molecular feature representations using the BAN enables the model to uncover intricate molecular interactions that traditional low-dimensional descriptors or predefined feature sets fail to detect. Furthermore, SCOPE's novel interaction module particularly strengthens its predictive capability within semi-inductive prediction scenarios, solidifying its potential as a powerful and broadly applicable tool in computational drug discovery.

\subsection*{Ablation study}

\begin{table}[ht]
\centering
\caption{\textbf{Ablation study on the filtered SCOPE dataset (averaged over five random runs).} The first three models evaluate the compound embedding design, followed by two models assessing the protein embedding design. The impact of the Bilinear Attention Network (BAN) layer is shown in the subsequent model. The final model integrates all components of the SCOPE design. (\textbf{Best}, \underline{Second Best})}
\begin{tabular}{l l l c c c}
\toprule
\textbf{Protein Encoding} & \textbf{Compound Encoding} & \textbf{Backbone} & \textbf{AUROC} & \textbf{AUPRC} & \textbf{F1} \\
\midrule
3D Graph HGNN & 1D Fingerprint & BAN+MLP & 0.829±0.012& 0.841±0.015& 0.771±0.019
\\
3D Graph HGNN & 2D Graph & BAN+MLP & 0.859±0.015& 0.873±0.015& 0.785±0.016
\\
3D Graph HGNN & 3D Graph GVP no Pooling & BAN+MLP & 0.860±0.014& \underline{0.874±0.014}& 0.786±0.015
\\
\hdashline
1D Onehot & 3D Graph GVP & BAN+MLP & 0.827±0.016& 0.846±0.017& 0.758±0.012
\\
1D CNN & 3D Graph GVP & BAN+MLP & \underline{0.873±0.008} & 0.871±0.007& \textbf{0.806±0.010}
\\
\hdashline
3D Graph HGNN & 3D Graph GVP & MLP & 0.859±0.024& 0.857±0.025& 0.789±0.026
\\
\hdashline
\rowcolor{gray!10} 3D Graph HGNN & 3D Graph GVP & BAN+MLP & \textbf{0.875±0.014} & \textbf{0.888±0.013} & \underline{0.803±0.016}
\\
\bottomrule
\end{tabular}
\label{table:ablation}
\end{table}

We conducted an ablation study to assess the individual contributions of protein encoding, compound encoding, and backbone design to the overall performance of the SCOPE model. Table~\ref{table:ablation} summarizes the key findings, highlighting the critical role each module plays in the model's effectiveness, with detailed experimental results provided in Supplementary Table 7.

To evaluate the impact of compound encoding using the 3D Graph GVP, we constructed three variants of the SCOPE model, each employing different compound encoding methods: 1D fingerprint~\cite{RDKitOpensourceCheminformatics}, 2D Graph~\cite{baiInterpretableBilinearAttention2023}, and 3D Graph GVP without global pooling, where the maximum atom size was set to 300. The results demonstrate that increasing the representational dimension of the compounds led to a significant improvement in model performance. Notably, the inclusion of pooled molecular encoding resulted in an over 1\% performance gain, highlighting the critical role of pooling techniques in enhancing molecular representations.

For protein encoding, we replaced the protein encoding block in the SCOPE model with two alternative methods: 1D One-hot encoding and 1D CNN~\cite{baiInterpretableBilinearAttention2023}. The results reveal that the 1D CNN-based encoding performs comparably to the 3D design and even outperforms it in terms of the F1 score. However, the 3D Graph HGNN architecture still outperforms the 1D CNN approach in AUROC and AUPRC, suggesting that the 3D design captures additional structural information that is important for model performance.

Finally, we examined the impact of the BAN layer within the backbone. Remarkably, removing the BAN layer caused a substantial drop in performance, bringing it to levels comparable to DrugBAN, which utilizes 1D CNN and 2D Graph for protein and compound encoding. This underscores the significance of both the encoding strategies and the backbone design in achieving optimal model performance.

In summary, the ablation study confirms the critical contributions of each module in the SCOPE model, validating the importance of its overall design and its role in improving performance.

\subsection*{Web server and database}

To enhance accessibility and usability, we have developed a web server and database for the SCOPE framework, accessible at \url{https://awi.cuhk.edu.cn/SCOPE/}. The platform allows users to input a SMILES string of a compound to perform search or prediction tasks. In search mode, the server identifies and returns all compounds from the SCOPE dataset with a structural similarity greater than 0.9 to the input molecule, calculated using RDKit. In prediction mode, the input molecule is paired with all proteins in the SCOPE library, providing interaction predictions with semi-inductive accuracy. Additionally, the complete SCOPE dataset is available for direct download from the website. Detailed information about the web development process and user instructions can be found in Supplementary Note and Supplementary Figure 3.

\begin{figure*}[htbp] 
    \begin{center} 
    \includegraphics[width=1\textwidth]{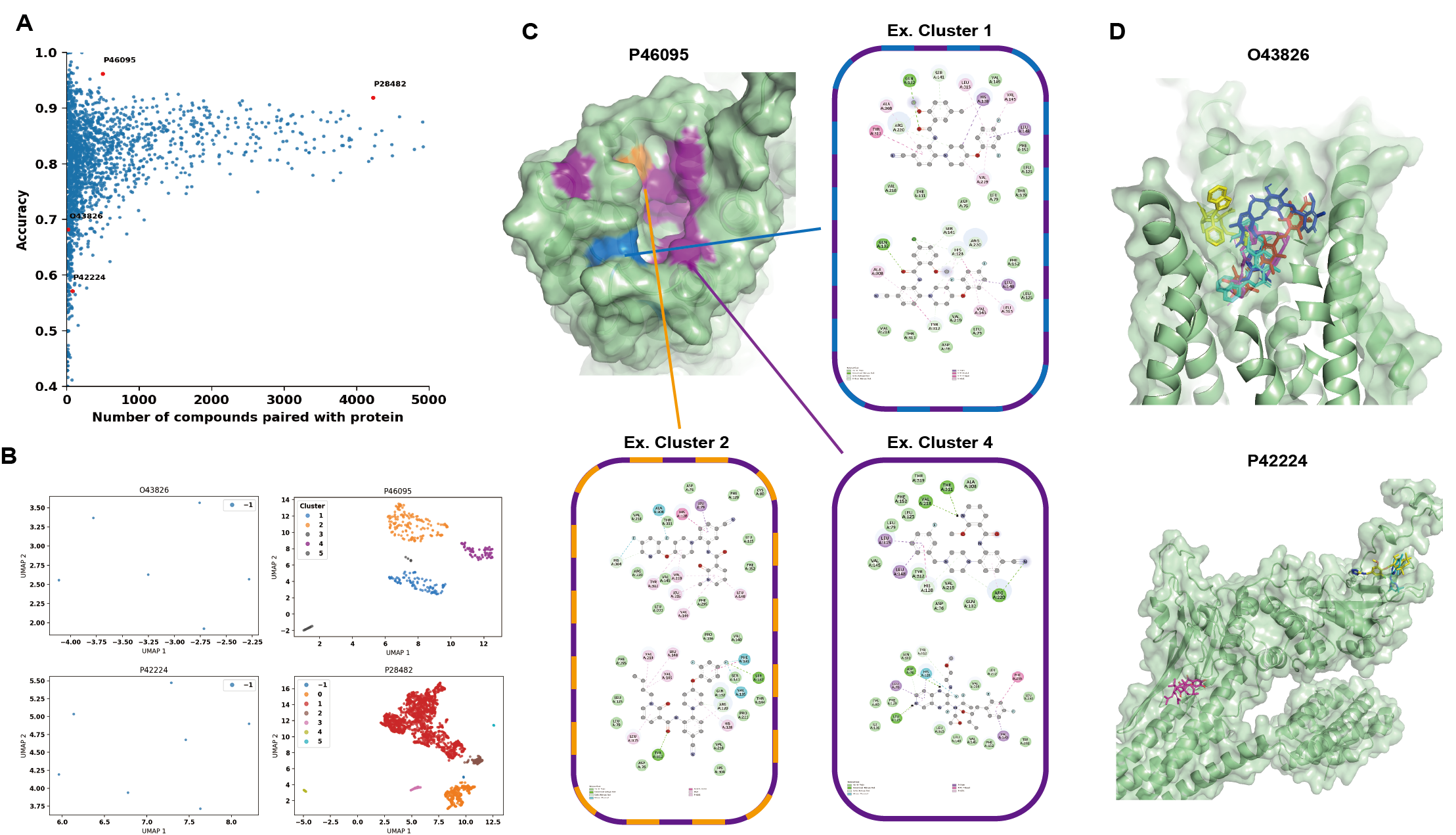} 
    \end{center} \vspace{-2em} 
    
    \caption{\textbf{Interpretability analysis of the model.}
    \textbf{(A)} Relationship between predictive performance and the number of known interactions per protein. As more known interactions are identified, accuracy stabilizes around 0.8–0.9, whereas fewer known interactions lead to greater variability.
    \textbf{(B)} UMAP-OPTICS clustering of BAN attention vectors for four representative proteins: two with robust predictive performance (P46095 and P28482) and two with lower accuracy (O43826 and P42224). Proteins enriched with interaction data (P46095 and P28482) exhibit clear clustering, suggesting that the model captures common binding features. In contrast, proteins with limited data (O43826 and P42224) do not show discernible clusters.
    \textbf{(C)} Docking analysis of ligands binding to P46095, highlighting shared and unique interaction residues across distinct ligand clusters. Unique critical residues are marked in blue and yellow, and universally important residues are shown in purple, indicating both global and cluster-specific binding determinants.
    \textbf{(D)} Analysis of proteins with relatively lower predictive accuracy. Although O43826 achieves an accuracy of about 0.7, its five known ligands—despite targeting the same pocket—adopt highly divergent poses, impeding a clear consensus. In the case of P42224 (accuracy <~0.6), the small number of known ligands not only bind in differing orientations but also occupy distinct pockets, challenging the model’s predictive capability.}
    \label{fig:inter}
    \end{figure*}
    
\subsection*{Model Interpretability}

As shown in Figure~\ref{fig:inter}A, prediction accuracy improves progressively with an increasing number of recorded protein interactions, highlighting the critical importance of sufficient drug–target interaction (DTI) data for robust model training. Proteins with limited known interactions exhibit considerable variability in predictive performance, whereas accuracy stabilizes at approximately 0.8–0.9 as interaction data expands.

To further explore this relationship, we examined four representative proteins (Figure~\ref{fig:inter}B). Proteins O43826 and P42224, characterized by fewer known interactions, displayed lower predictive accuracy. In contrast, proteins P46095 and P28482, benefiting from richer datasets, achieved consistently higher accuracy. Interaction embeddings derived from the BAN attention module were analyzed using UMAP \cite{mcinnes2018umap} for dimensionality reduction and OPTICS \cite{ankerst1999optics} for clustering. Proteins with extensive interaction data (P46095 and P28482) demonstrated clear, distinct clusters, indicating that the model effectively captures meaningful binding features. Conversely, sparse interaction data for O43826 and P42224 resulted in indistinct or no observable clusters, reflecting difficulties in recognizing consistent binding patterns.

A deeper analysis of docking results for the GPCR protein P46095 (Figure~\ref{fig:inter}C) revealed that clusters identified by the model correspond to chemically distinct binding modes within the canonical binding pocket. Specifically, interactions in clusters 1, 2, and 4 commonly involved residues 79, 128, 132, 145, 219, 220, 312, and 315, while unique residues 295 and 304 characterized cluster 1, and residue 141 was distinctive for cluster 2. These findings confirm that our model captures nuanced and chemically relevant features governing ligand specificity.

Further examination of proteins with relatively low predictive accuracy (Figure~\ref{fig:inter}D) revealed additional complexity. Protein O43826, despite an accuracy of around 0.7, had five known ligands adopting significantly different binding poses within the same pocket, complicating the identification of consistent patterns. Similarly, protein P42224 (accuracy <~0.6) had limited ligands binding in various orientations and distinct pockets, presenting significant modeling challenges.

Collectively, these analyses emphasize the necessity of sufficient and diverse interaction data to enable accurate predictions of drug–target interactions. They also illustrate the interpretative capability of our model, highlighting its ability to identify both global and ligand-specific binding features critical for understanding molecular interactions. For further methodological details and additional analyses, see the Supplementary Note.

\subsection*{Efficient Target Discovery for Ginsenoside Rh1 Anti-Cancer Activity Using SCOPE-DTI}

\begin{figure*}[!t]
    \begin{center}
    \includegraphics[width=0.9\textwidth]{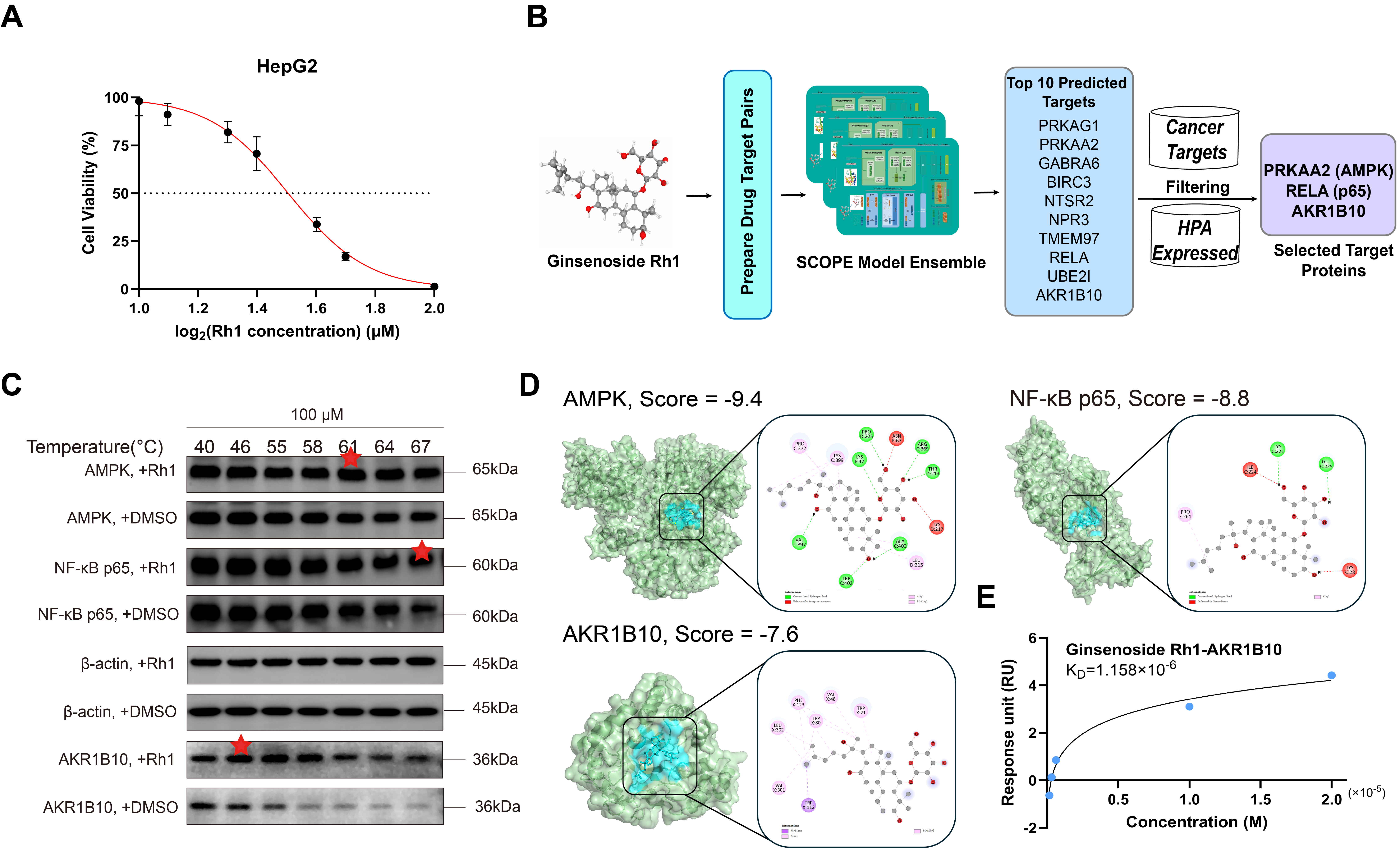}
    \end{center}
    \vspace{-2em}
    \caption{\textbf{Experimental validation of anti-cancer targets of Ginsenoside Rh1 predicted by SCOPE-DTI in HepG2 cells.} 
\textbf{(A)} Dose-dependent cytotoxicity of Ginsenoside Rh1 in HepG2 cells. Cell viability was assessed after 24-hour treatment using the CCK-8 assay. Data are normalized to untreated controls and presented as mean~±~SD. 
\textbf{(B)} Target identification of Ginsenoside Rh1 in HepG2 cells, highlighting AMPK, NF-$\kappa$B p65, and AKR1B10 as candidate proteins for experimental validation based on their expression levels and functional relevance. 
\textbf{(C)} Validation of Rh1 binding to AMPK, NF-$\kappa$B p65, and AKR1B10 using the cellular thermal shift assay (CETSA) combined with Western blot analysis. Lysates from Rh1- or DMSO-treated HepG2 cells were subjected to thermal gradients, and the thermal stability of each target protein was analyzed. The red star indicates the initial temperature at which a difference between the drug-treated group and the control group was observed.
\textbf{(D)} Predicted binding conformations of Rh1 within the target proteins based on molecular docking simulations. Enlarged views illustrate the binding pockets and critical interactions that support Rh1's affinity for the predicted targets. 
\textbf{(E)} Surface plasmon resonance (SPR) analysis of Rh1 binding to AKR1B10. The dissociation equilibrium constant (KD) was determined to quantify the binding affinity between Rh1 and AKR1B10.}
    \label{fig:case}
\end{figure*}

To demonstrate the practical applicability of the SCOPE framework, we selected an active compound with unknown targets, aiming to computationally predict its molecular targets using SCOPE and validate these predictions experimentally. Ginsenoside saponins, a class of bioactive compounds found abundantly in Panax species, are widely used in both traditional and modern medicine due to their diverse pharmacological effects, benefiting millions globally \cite{tangPanaxNotoginsengAlleviates2023}. Among these, Ginsenoside Rh1 is recognized as a major final metabolite in humans \cite{qiMetabolismGinsengIts2011, tamGinsenosideRh1Systematic2018}. Although Rh1 has demonstrated anticancer activity in several contexts, its potential against liver cancer remains unexplored, and its specific molecular targets have not been identified, limiting its further development and therapeutic application \cite{lyuGinsenosideRh1Inhibits2019, jungProtopanaxatriolGinsenosideRh12013}. In this study, we first evaluated the cytotoxic effects of Rh1 on HepG2 cells using the CCK-8 assay.  As shown in Figure~\ref{fig:case}A and Supplementary Table 8, the IC$_{10}$, IC$_{30}$, and IC$_{50}$ values of Rh1 were determined to be 16.55~\(\mu M\), 24.86~\(\mu M\), and 32.04~\(\mu M\), respectively, confirming that Rh1 significantly inhibits HepG2 cell proliferation.

Having established its inhibitory effect, we next utilized the SCOPE framework to predict potential anti-cancer targets of Rh1 in HepG2 cells. As shown in Figure~\ref{fig:case}B, Rh1 was computationally paired with all candidate targets in the SCOPE database, and multiple models trained with different random parameters were employed to rank these targets. After averaging the predictive scores across all models, we extracted the top 10 ranked proteins. Two filtering criteria were applied to refine the results: (1) exclusion of targets with no reported anti-cancer relevance according to prior research \cite{savagePancancerProteogenomicsExpands2024}, and (2) removal of targets not expressed in HepG2 cells based on the Human Protein Atlas (HPA) \cite{uhlenProteomicsTissuebasedMap2015}. This refinement process identified three high-confidence targets: AMPK, NF-\(\kappa\)B p65, and AKR1B10, which were selected for further experimental validation.

To experimentally confirm these computational predictions, we employed the cellular thermal shift assay (CETSA) combined with Western blot (CETSA-WB) analysis. HepG2 cells were treated with Rh1 or DMSO and exposed to a temperature gradient ranging from 40\(^\circ\)C to 67\(^\circ\)C. As shown in Figure~\ref{fig:case}C, Rh1 treatment significantly enhanced the thermal stability of AKR1B10 compared to control conditions, while also imparting moderate stabilization to AMPK and NF-\(\kappa\)B p65. To gain deeper insight into the molecular basis of these interactions, molecular docking simulations were performed (Figure~\ref{fig:case}D). The docking results indicate that Rh1 occupies the ideal binding pocket of AKR1B10, likely hindering the attachment of its cofactor (e.g., NADPH) and impairing its enzymatic function. Moreover, Rh1’s interaction with the AMP-binding site of AMPK may disrupt AMP-mediated regulation, thus affecting AMPK activation and downstream signaling. Additionally, Rh1’s binding to the DNA-binding interface of NF-\(\kappa\)B p65 could block DNA association, reducing the transcriptional activity of this key regulator.

Among these targets, AKR1B10 exhibited the most pronounced stabilization in the CETSA-WB assay, prompting further investigation into the direct binding between Rh1 and AKR1B10 using surface plasmon resonance (SPR). As shown in Figure~\ref{fig:case}E, SPR analysis confirmed a direct, high-affinity binding interaction between Rh1 and AKR1B10, with a dissociation equilibrium constant ( \( K_\mathrm{D} \) ) of 1.158~\(\times 10^{-6}\)~M. Collectively, these results establish AKR1B10 as a key molecular target of Rh1, elucidating its mechanistic role in the observed anti-cancer effects in HepG2 cells and further validating the utility and accessibility of the SCOPE framework for target identification.

\section*{Conclusion}

In this study, we introduce SCOPE-DTI, a novel framework developed to enhance the practical utility of deep learning-based drug–target interaction (DTI) prediction. By constructing the largest semi-inductive human DTI dataset to date, we provide a solid foundation for further investigation in this domain. Our approach incorporates target-level data balancing, which ensures a more reliable assessment of model performance under real-world conditions. By leveraging three-dimensional structural representations of both proteins and compounds, combined with an attention-based model architecture, SCOPE-DTI demonstrated superior predictive performance across multiple benchmark datasets. Furthermore, we validated its structural interpretability, underscoring its potential for offering mechanistic insights into drug–target interactions. In addition to these technical advancements, SCOPE-DTI's practical applicability was exemplified by the experimental identification of anticancer targets of Ginsenoside Rh1. To foster broader adoption, we also developed a user-friendly web interface, providing seamless access to both our database and model for drug discovery research.

Despite these accomplishments, several areas remain for future improvement. Our ablation studies indicate that incorporating 3D compound structures results in more significant performance gains than integrating 3D protein structures. This discrepancy likely reflects the current limitations of AlphaFold2 in generating highly accurate protein representations. Future iterations, such as AlphaFold3 or more refined encoding techniques, could provide more accurate protein structure predictions, thereby enabling more precise drug-target interaction predictions. Furthermore, while the bilinear attention mechanism effectively captures residue-level interaction patterns, our case analyses indicate that its interpretability currently remains meaningful primarily at the cluster level rather than at an individual, residue-specific level, as the attention vectors do not consistently align directly with observed interactions on a case-by-case basis. Incorporating more granular data, such as detailed pocket-level binding information, along with targeted model refinements, could further enhance both the interpretability and accuracy of the framework.

\section*{Methods}

\subsection*{SCOPE Dataset Development}

We developed the SCOPE dataset as a comprehensive, multi-source, and well-annotated resource tailored for drug-target interaction (DTI) prediction. This dataset integrates DTI data from a wide range of public sources, including ChEMBL activity\cite{zdrazilChEMBLDatabase20232024}, PubChem activity\cite{kimPubChem2023Update2023}, DrugBank\cite{knoxDrugBank60DrugBank2024}, BindingDB\cite{gilsonBindingDB2015Public2016}, DrugCentral\cite{avramDrugCentral2023Extends2023}, TTD\cite{zhouTTDTherapeuticTarget2024}, Pharos\cite{kelleherPharos2023Integrated2023}, PROMISCUOUS\cite{galloPROMISCUOUS20Resource2021}, GtoPdb\cite{hardingIUPHARBPSGuide2024}, Human\cite{liuImprovingCompoundproteinInteraction2015}, BioSNAP\cite{leskovecSNAPGeneralPurposeNetwork2016}, KIBA\cite{tangMakingSenseLargescale2014}, and DAVIS\cite{davisComprehensiveAnalysisKinase2011}. Additionally, these sources encompass more than 20 reference data sources. However, the raw data presented several challenges: (1) lack of unified annotations, (2) varying scales and pharmacological evaluation metrics, and (3) inconsistent data quality with potential erroneous entries. To address these issues, we performed comprehensive data cleaning and standardization.

\textbf{Protein and Compound Annotation.} We unified protein annotations by Uniprot database and filtered all human proteins, focusing on druggable targets for interaction prediction with small molecules. Proteins were classified into pharmacological target families (e.g., G protein–coupled receptors (GPCRs), kinases) using the IUPHAR database. To provide structural insights, we generated three-dimensional (3D) structures for all proteins using AlphaFold2. Compounds were annotated using identifiers from the PubChem and ChEMBL databases. We generated 3D structures of the compounds using RDKit and optimized their conformations with the Merck Molecular Force Field (MMFF).

\textbf{Interaction Label Unification.} To harmonize different scales and pharmacological evaluation metrics across various data sources, we assigned interaction labels based on PubChem activity distributions, classifying them as either positive (1) or negative (0). Specific classification criteria are detailed in Supplementary Note. This approach allowed us to standardize experimental information from different sources into a consistent binary classification.

\textbf{Multi-Source Data Integration.} To further improve data quality, we consolidated multiple data sources by retaining only interactions consistently labeled as positive across all sources. If any source labeled an interaction as negative, we considered the interaction negative in our dataset. This conservative approach maximized the reliability of positive samples.

\textbf{Mitigating Target-Level Interaction Imbalance.} To address target-level interaction imbalance -- where a protein \( p_i \) exhibits a disproportionate number of interactions from one class -- we implemented a dataset filtering strategy. For each protein \( p_i \), we consider its interaction set \( \mathcal{I}(p_i) = \{ (p_i, c_j, l_{ij}) \} \),where \( c_j \) denotes a compound and \( l_{ij} \in \{0, 1\} \) indicates the interaction label. The imbalance is quantified through two counters: \( N_0 \)(negative interactions) and \( N_1 \) (positive interactions) within \( \mathcal{I}(p_i) \).

If \( p_i \) had more than 75\% of its interactions belonging to one class, we performed stratified random sampling to obtain a subset \( \mathcal{I}'(p_i) \) that satisfies:

\begin{equation}
\frac{\min(N_0', N_1')}{N_0' + N_1'} \geq 0.25,
\label{eq:class_balance}
\end{equation}

where \( N_0' \) and \( N_1' \) are the counts of negative and positive interactions in \( \mathcal{I}'(p_i) \), respectively. This ensures that the minority class represents at least 25\% of the interactions for \( p_i \), achieving a class ratio between 25\% and 75\%. Proteins with insufficient interactions after filtering were excluded to maintain data quality.

\textbf{Visualization of Data Cleaning Effects.} To demonstrate the impact of our data cleaning and balancing procedures, we analyzed the distribution of interaction ratios per protein before and after filtering. For each protein \( p_i \), we calculated the ratio of positive interactions:

\begin{equation}
R(p_i) = \frac{N_1}{N_0 + N_1},
\end{equation}

where \( N_0 \) and \( N_1 \) are the counts of negative and positive interactions, respectively, for protein \( p_i \). We grouped proteins into intervals based on \( R(p_i) \) with a specified interval length (e.g., 0.02) and plotted the number of proteins within each interval. This visualization highlighted the reduction of proteins with extreme interaction imbalances after filtering.

\textbf{Dataset Statistics and Comparative Analysis.} We quantified the SCOPE dataset by calculating the numbers of unique compounds, targets, and total interactions. Our dataset encompasses a significantly larger number of compounds and interactions compared to existing datasets, both before and after the filtering procedures. To illustrate the scale and comprehensiveness of SCOPE, we generated plots that compare the data volume with that of previous datasets, highlighting our dataset's superiority in size and diversity. Detailed statistical information and comparative analyses are provided in Supplementary Note.

\subsection*{SCOPE Framework Architecture}

\textbf{HGNN for Protein Structure.} A protein \( P_i \in \mathcal{P} \), composed of \( M \) amino acid residues, can be represented by its primary sequence \( S_i = (v_1, v_2, \dots, v_M) \), where each residue \( v_m \) belongs to one of the 20 standard amino acid types. In their physicochemical environment, these sequences fold into stable three-dimensional (3D) structures\cite{zhangProteinRepresentationLearning2023}. To capture both the sequence and structural information of protein \( P_i \), we construct a residue-level heterogeneous protein graph \( G_P^{(i)} = (V_P^{(i)}, E_P^{(i)}, R) \), where \( V_P^{(i)} \) represents the residues as nodes, \( E_P^{(i)} \) denotes the edges connecting them, and \( R \) specifies the edge types. 

In this work, we use two types of edges to represent the relationships between residues. Sequential edges connect residues that are adjacent in the primary sequence, preserving the natural order of the amino acids. Radius edges connect residues whose spatial Euclidean distance, calculated based on the geometric centers of all their atoms, is below a predefined threshold \( d_r \). This approach ensures that both local sequence context and spatial proximity within the folded protein are encoded.

To encode the heterogeneous protein graph \( G_P = (V_P, E_P, R) \), we employ a Heterogeneous Graph Neural Network (HGNN)\cite{zhangHeterogeneousGraphNeural2019}. Each edge type \( r \in R \) is associated with a weight matrix \( W_r^{(l)} \) at layer \( l \), and a shared weight matrix \( W_h^{(l)} \) is used to combine the aggregated messages from different edge types. 

The node embeddings are updated iteratively across \( L \) layers. At layer \( l \) (where \( 1 \leq l \leq L \)), the embedding of each residue \( v_m \) is computed as:

\begin{equation}
h_m^{(l)} = \text{BN} \left( \text{ReLU} \left( W_h^{(l)} \cdot \sum_{r \in R} \sum_{v_n \in \mathcal{N}_r(m)} W_r^{(l)} h_n^{(l-1)} \right) \right),
\label{eq:hgnn_update}
\end{equation}

where \( h_m^{(0)} = x_m \) represents the initial input feature of residue \( v_m \), and \( h_m^{(l)} \) is the embedding at layer \( l \). The neighbors \( \mathcal{N}_r(m) \) are defined as residues connected to \( v_m \) by edge type \( r \). The batch normalization function \( \text{BN}(\cdot) \) and rectified linear unit activation \( \text{ReLU}(\cdot) \) are applied for stable training, while \( W_r^{(l)} \in \mathbb{R}^{d \times d} \) and \( W_h^{(l)} \in \mathbb{R}^{d \times d} \) are learnable weight matrices. This formulation ensures that the heterogeneous nature of the graph is effectively captured, integrating sequence and structural information into a unified representation.

This formulation ensures that messages from different edge types are appropriately transformed and aggregated.

\textbf{GVP for Drug Encoding.} Given the three-dimensional (3D) coordinates of atoms in a molecule, we represent the molecular structure as a graph \( G_d = (V_d, E_d) \), where the nodes \( V_d \) correspond to the atoms of the molecule, and the edges \( E_d \) are defined between pairs of atoms whose Euclidean distance is less than \( 4.5\,\text{\AA} \)\cite{townshendATOM3DTasksMolecules2022}. This representation captures both the chemical and spatial relationships within the molecule.

\textit{Node Features:} For each atom \( i \), we construct a node feature \( h_i^{(0)} = (h_i^{(0),v}, h_i^{(0),s}) \), which consists of both a vector component \( h_i^{(0),v} \) and a scalar component \( h_i^{(0),s} \). The vector feature \( h_i^{(0),v} = c_i \in \mathbb{R}^3 \) represents the atom's 3D coordinates in space. The scalar feature \( h_i^{(0),s} \in \mathbb{R}^{74} \) is a 74-dimensional integer vector that describes the atom with eight types of information: the atom type, the atom degree, the number of implicit hydrogens, the formal charge, the number of radical electrons, the atom hybridization, the number of total hydrogens, and whether the atom is aromatic\cite{liDGLLifeSciOpenSourceToolkit2021}.

\textit{Edge Features:} For each edge \( (i, j) \in E_d \), we define an edge feature \( e_{ji} = (e_{ji}^v, e_{ji}^s) \). The vector feature \( e_{ji}^v = \frac{c_j - c_i}{\| c_j - c_i \|} \in \mathbb{R}^3 \) is the unit vector pointing from atom \( i \) to atom \( j \). The scalar feature \( e_{ji}^s = \text{RBF}(\| c_j - c_i \|) \in \mathbb{R}^{16} \) encodes the pairwise distance using 16 Gaussian radial basis functions (RBFs) with centers evenly spaced between 0 and \( 4.5\,\text{\AA} \).

\textit{Molecular Graph Neural Network:} To learn a representation for the input molecule, we utilize a Graph Neural Network based on Geometric Vector Perceptrons (GVPs)\cite{jingLearningProteinStructure2021}. Each node embedding \( h_i^{(l)} = (h_i^{(l),v}, h_i^{(l),s}) \) consists of vector and scalar components.

At each layer $l$, the node embeddings are updated using the following equation:

\begin{equation}
h_i^{(l)} = h_i^{(l-1)} + \text{GVP} \left( h_i^{(l-1)}, \sum_{j \in \mathcal{N}(i)} \text{GVP} \left( h_j^{(l-1)}, e_{ji} \right) \right),
\end{equation}

where $\mathcal{N}(i)$ denotes the set of neighboring nodes of node $i$, $\text{GVP}(\cdot, \cdot)$ represents a GVP layer that processes node and edge features, and $h_i^{(0)} = v_i$ is the initial feature of node $i$. This update rule incorporates both the features of neighboring atoms and the geometric information from edge features.

After $L$ layers of message passing, we obtain the final node embeddings $h_i^{(L)}$ for all nodes in the graph. To derive a fixed-size representation of the entire molecule, we apply a global add pooling operation over all node embeddings:

\begin{equation}
h_d = \sum_{i \in V_d} h_i^{(L)},
\end{equation}

where $h_d \in \mathbb{R}^{d}$ is the learned representation of the input molecule.

\textbf{Pairwise Interaction Learning.} To capture pairwise local interactions between drugs and proteins, we employ a bilinear attention network module comprising two key layers: a bilinear interaction map to compute pairwise attention weights, and a bilinear pooling layer over the interaction map to extract a joint drug–protein representation.

Given the hidden representations from the encoders---\(H_d \in \mathbb{R}^{N \times D_d}\) for the drug and \(H_p \in \mathbb{R}^{M \times D_p}\) for the protein---we compute the pairwise interaction matrix \(I \in \mathbb{R}^{N \times M}\) as:
\begin{equation}
I = \bigl(\sigma\bigl(H_d U\bigr)\,q^\top\bigr)\,\odot\,\sigma\bigl(H_p V\bigr)^\top,
\end{equation}
where \(U \in \mathbb{R}^{D_d \times K}\) and \(V \in \mathbb{R}^{D_p \times K}\) are learnable weight matrices, \(q \in \mathbb{R}^{K}\) is a learnable weight vector, \(\sigma(\cdot)\) is an activation function (e.g., sigmoid), \(\odot\) denotes element-wise multiplication (Hadamard product), and the superscript \(\top\) indicates matrix transpose. Each element \(I_{i,j}\) represents the interaction score between the \(i\)-th atom of the drug and the \(j\)-th residue of the protein.

For intuitive understanding, an element \(I_{i,j}\) can be expressed as:
\begin{equation}
I_{i,j} = q^\top \Bigl(\sigma\bigl(U^\top h_d^i\bigr)\,\odot\,\sigma\bigl(V^\top h_p^j\bigr)\Bigr),
\end{equation}
where \(h_d^i\) and \(h_p^j\) are the embeddings of the \(i\)-th atom and the \(j\)-th residue, respectively.

The joint representation \(f' \in \mathbb{R}^{K}\) is then computed via bilinear pooling:
\begin{equation}
f' = \Bigl(\sigma\bigl(H_d U\bigr)^\top\,I\,\sigma\bigl(H_p V\bigr)\Bigr)\,1,
\end{equation}
where \(1 \in \mathbb{R}^{M}\) is a vector of ones used for summation over the protein residues. To reduce dimensionality, we apply sum pooling:
\begin{equation}
f = \text{SumPool}\bigl(f', s\bigr),
\end{equation}
resulting in the final feature vector \(f \in \mathbb{R}^{K/s}\).

To compute the interaction probability, we feed the joint representation \(f\) into a decoder consisting of a fully connected layer followed by a sigmoid activation function:
\begin{equation}
p = \sigma\bigl(W_o f + b_o\bigr),
\end{equation}
where \(W_o\) and \(b_o\) are learnable weight parameters.

Finally, we jointly optimize all learnable parameters using backpropagation. The training objective is to minimize the cross-entropy loss with L2 regularization:
\begin{equation}
\mathcal{L} = -\sum_{i} \Bigl[y_i \log p_i + (1 - y_i) \log (1 - p_i)\Bigr] \;+\; \frac{\lambda}{2}\|\theta\|_2^2,
\end{equation}
where \(\theta\) represents the set of all learnable parameters, \(y_i\) is the ground-truth label for the \(i\)-th drug--protein pair, \(p_i\) is the predicted probability, and \(\lambda\) is the regularization hyperparameter. Overall, this pairwise interaction learning approach is highly inspired by and adapted from DrugBAN~\cite{baiInterpretableBilinearAttention2023}.

\subsection*{Experimental setting}

\textbf{Datasets.} We evaluated the SCOPE model and five state-of-the-art baseline models on four public DTI datasets: BindingDB, KIBA, Human, and our newly constructed SCOPE dataset. The SCOPE dataset includes various protein families annotated by IUPHAR, such as G-protein-coupled receptors (GPCRs), kinases, ion channels, and nuclear hormone receptors, along with our total dataset. All datasets were tested in both their original and debiased versions. The BindingDB dataset \cite{gilsonBindingDB2015Public2016} is a web-accessible database of experimentally validated binding affinities between small drug-like molecules and proteins; we utilized a low-bias version of this dataset \cite{baiHierarchicalClusteringSplit2021}. The KIBA dataset \cite{tangMakingSenseLargescale2014} integrates various bioactivity measurements to provide a comprehensive set of drug–target interactions, focusing particularly on kinase inhibitors. The Human dataset, constructed by Liu et~al.\ \cite{liuImprovingCompoundproteinInteraction2015}, includes highly credible negative samples generated via an \textit{in~silico} screening method; following previous studies \cite{chenTransformerCPIImprovingCompoundprotein2020,zhengPredictingDrugProtein2020}, we used the balanced version containing an equal number of positive and negative samples.

\textbf{Implementation.} SCOPE model is implemented in Python 3.9 and PyTorch 2.2.0 \cite{paszkePyTorchImperativeStyle2019}, along with functions from PyG 2.5.2 \cite{feyFastGraphRepresentation2019} DGL 2.2.5 \cite{wangDeepGraphLibrary2020}, DGLlifeSci 0.3.2 \cite{liDGLLifeSciOpenSourceToolkit2021}, Scikit-learn 1.0.2 \cite{pedregosaScikitlearnMachineLearning2018}, Numpy 1.20.2 \cite{harrisArrayProgrammingNumPy2020}, Pandas 1.5.2 \cite{mckinneyDataStructuresStatistical2010} and RDKit 2021.03.2 \cite{RDKitOpensourceCheminformatics}. The batch size is set to be 64 and the Adam optimizer is used with a learning rate of 5e-5. We allow the model to run for at most 100 epochs for all datasets. The best performing model is selected at the epoch giving the best AUROC score on the validation set, which is then used to evaluate the final performance on the test set. 
The protein encoder HGNN utilizes an embedding dimension of 320 and processes protein sequences with a maximum allowed length of 2000 amino acids. To construct the protein 3D graph, we set an edge cutoff distance of 10\,\AA{}. The HGNN comprises 4 layers, and we include a fully connected layer with bias to enhance model capacity. For the drug feature encoder, the atom input dimensions are set to $[74,\,1]$, capturing both scalar and vector features. The atom hidden dimensions are $[320,\,64]$, allowing the model to learn complex representations. Similarly, the edge input dimensions are $[16,\,1]$, and the edge hidden dimensions are $[32,\,1]$. The GVP model consists of 3 layers and incorporates a dropout rate of 0.1 to prevent overfitting. An edge cutoff distance of 4.5\,\AA{} is used, and we compute 16 radial distribution functions to capture spatial relationships between atoms. Notably, there is no maximum length restriction for compounds in our model. In the bilinear attention module, we employ two attention heads to enhance interpretability while capturing intricate interactions between drugs and proteins. The latent embedding size is set to 768, and we use a sum pooling window size of 3 to aggregate features effectively. The decoder is a fully connected network with 512 hidden neurons, enabling the model to make accurate predictions based on the learned representations.

\textbf{Baselines.} We compare the SCOPE model with the following five methods for DTI prediction: (1) \textbf{SVM}, a shallow machine learning algorithm; (2) \textbf{MLP}, a simple deep neural network with hidden dimensions [2048, 512, 128, 32], applied to the concatenated ECFP4 and PSC fingerprint features; (3) \textbf{MolTrans}, a deep learning model that uses the transformer architecture to encode drug and protein features, with a CNN-based module to capture sub-structural interactions; (4) \textbf{TransformerCPI}, a Transformer-based model with an encoder for protein sequences and a decoder for molecular graphs, leveraging multi-head attention to extract interaction features; (5) \textbf{DrugBAN}, a model that encodes drug molecules and protein sequences using graph convolutional networks and 1D-convolutional neural networks, followed by a bilinear attention network to capture pairwise interactions and a fully connected decoder for DTI prediction.  For the above deep DTI models, we follow the recommended model hyper-parameter settings described in their original papers.

\subsection*{Experimental validation of anti-cancer targets of Ginsenoside Rh1}
A concise version of the experimental methods is provided here, with detailed descriptions available in Supplementary Note.

\textbf{Cell culture.} HepG2 hepatocellular carcinoma cells were cultured in Minimum Essential Medium (Gibco, cat.no.310950251) supplemented with 10\% fetal bovine serum (Gibco, cat.no.26170043), 1× non-essential amino acids (Gibco, cat.no. 11140050), 1× GlutaMAX (Gibco, cat.no.35050061), and 1\% penicillin/streptomycin (Gibco, cat.no.15140122). Cells were maintained at 37◦C in a 5\% CO2 humidified incubator and passaged using Trypsin-EDTA (Gibco, cat.no.2585625). 

\textbf{Cell viability assay.} HepG2 cells were seeded at 1 × 104 cells/well in 96-well plates and allowed to adhere overnight. Cells were treated with ginsenoside Rh1 (10–100 µM) for 24 h, with untreated cells as controls. Cell viability was assessed by CCK-8 assay using Enhanced Cell Counting Kit-8 (Beyotime Biotechnology, cat.no.C0042), measuring optical density at 450 nm.

\textbf{CETSA-WB.} HepG2 lysates were incubated with 100 µM Rh1 or 1\% DMSO (v/v) at room temperature for 10 min and then heated at defined temperatures (40–67◦C) for 3 min. After cooling and centrifugation, supernatants were subjected to SDS-PAGE and transferred onto PVDF membranes. Membranes were probed with anti-AKR1B10 (Abcam, cat.no.Ab96417), anti- Anti-AMPK alpha 2 (Abcam, cat.no.EP20772), and anti-NF-\(\kappa\)B p65 (Abcam, cat.no.E379) antibodies, followed by chemiluminescent detection and imaging.

\textbf{Molecular docking.} The crystal structures of AKR1B10 (PDB: 4JII), AMPK (PDB: 4ZHX), and NF-\(\kappa\)B p65 (PDB: 1NFI), were obtained from the Protein Data Bank. Ginsenoside Rh1 was docked into these proteins using CBDOCK2. The top-scoring docking conformations were visualized and analyzed in PyMOL 3.1.3.

\textbf{Surface plasmon resonance (SPR).} SPR analysis was performed with AKR1B10 immobilized on an SPR chip. Rh1 solutions (200–1.5625~$\mu$M) were injected, and sensorgrams were recorded to determine the association ($k_{a}$), dissociation ($k_{d}$), and equilibrium dissociation ($K_{D}$) constants.

\textbf{Statistical analysis.} Data were analyzed using GraphPad Prism (v10.2.3) and are presented as mean $\pm$ SD. Statistical significance was defined as $p<0.05$.

\section*{Data availability}
The experimental data with each split strategy is available at \url{https://awi.cuhk.edu.cn/SCOPE/}. All data used in this work are from public resource.  The Human source is at \url{https://github.com/lifanchen-simm/transformerCPI}; The BindingDB source is at \url{https://github.com/peizhenbai/DrugBAN/tree/main/datasets} and the KIBA source is at \url{https://github.com/luoyunan/KDBNet}.

\section*{Code availability}
The source code and implementation details of SCOPE are freely available at GitHub repository (\url{https://github.com/Yigang-Chen/SCOPE-DTI}).

\section*{Additional information}
\textbf{Competing interests}: the authors declare no competing interests.

\bibliography{main}

\section*{Acknowledgements}

We would like to express our sincere gratitude to the Vincent \& Lily Woo Foundation for their generous support of the Vincent \& Lily Woo Fellowship in Memory of Dr Albert Wong. This fellowship, endowed by the Vincent \& Lily Woo Foundation, is provided through MCMIA Foundation Limited, and we are deeply grateful for their contribution to our research.

\section*{Author contributions statement}

Y.C., X.J., Z.Z. (Ziyue Zhang), H.-Y.H., and H.-D.H. conceived the study. Data collection was conducted by Y.C., Z.Z. (Ziyue Zhang), X.L., J.F., and Y.H. Y.C., Z.Z. (Ziyue Zhang), K.C., and A.W. designed the methodology. Computational analysis was performed by Y.C., Z.Z. (Ziyue Zhang), Y.Z. (Yuming Zhou), K.C., C.S., X.L., and Y.H., with interpretability validation contributed by Y.C., and Z.Z. (Ziyue Zhang). Wetlab experiments were executed by X.J., Y.L., Z.Z. (Zihao Zhu), Y.Z. (Yangyi Zhang), K.W., and Y.-C.-D.L. The original manuscript was drafted by Y.C., X.J., Z.Z. (Ziyue Zhang), C.S., H.-Y.H., S.-C.Y., Y.-C.-D.L., and H.-D.H. All authors reviewed and approved the final manuscript.

\end{document}